\documentclass[sigconf,natbib=true]{acmart}

\usepackage{multirow}

\AtBeginDocument{%
  }

\setcopyright{none}
\acmConference{SIGIR 2026 SynthIR Workshop}{}{}
\acmISBN{}
\acmDOI{}
\copyrightyear{}
\DeclareMathOperator*{\argmax}{argmax}

\newcommand{\us}{\_\allowbreak}

\makeatletter
\def\@ACM@checkaffil{}
\makeatother

\begin{document}

\title{Towards Vision-Free CIR: Attribute-Augmented Scoring and LLM-Based Reranking for Zero-Shot Composed Image Retrieval}

\author{Ryotaro Shimada}
\affiliation{
  \institution{The University of Tokyo}
}
\email{shimada@hal.t.u-tokyo.ac.jp}

\author{Yu-Chieh Lin}
\affiliation{
  \institution{Kioxia Corporation}
}
\email{yuchieh.lin@kioxia.com}

\author{Yuji Nozawa}
\affiliation{
  \institution{Kioxia Corporation}
}
\email{yuji1.nozawa@kioxia.com}

\author{Youyang Ng}
\affiliation{
  \institution{Kioxia Corporation}
}
\email{youyang.ng@kioxia.com}

\author{Osamu Torii}
\affiliation{
  \institution{Kioxia Corporation}
}
\email{osamu.torii@kioxia.com}

\author{Yusuke Matsui}
\affiliation{
  \institution{The University of Tokyo}
}
\email{matsui@hal.t.u-tokyo.ac.jp}

\begin{abstract}
Recent work has shown that ``Vision-Free'' approaches (representing images as text) can be effective for standard image retrieval tasks. However, it remains unclear whether this paradigm can effectively handle a more complex, multimodal task, Composed Image Retrieval (CIR), due to the inherent information loss in textual descriptions. In this paper, we introduce a Vision-Free CIR framework that addresses this challenge through two key techniques: (1) Attribute-Augmented Hybrid Scoring, which compensates for lost visual details via explicit attribute matching, and (2) LLM-Based Reranking, which verifies semantic consistency of top candidates. Experiments on the open-domain CIRR dataset show that our approach outperforms existing Zero-shot CIR methods (44.04\% R@1, +8.79\%). On FashionIQ, our results highlight the trade-off between semantic reasoning and fine-grained visual matching. Ablation studies reveal that both attribute-augmented scoring and LLM-Based Reranking consistently improve performance.
\end{abstract}



\maketitle

\begin{figure*}[t]
    \centering
    \includegraphics[width=\linewidth]{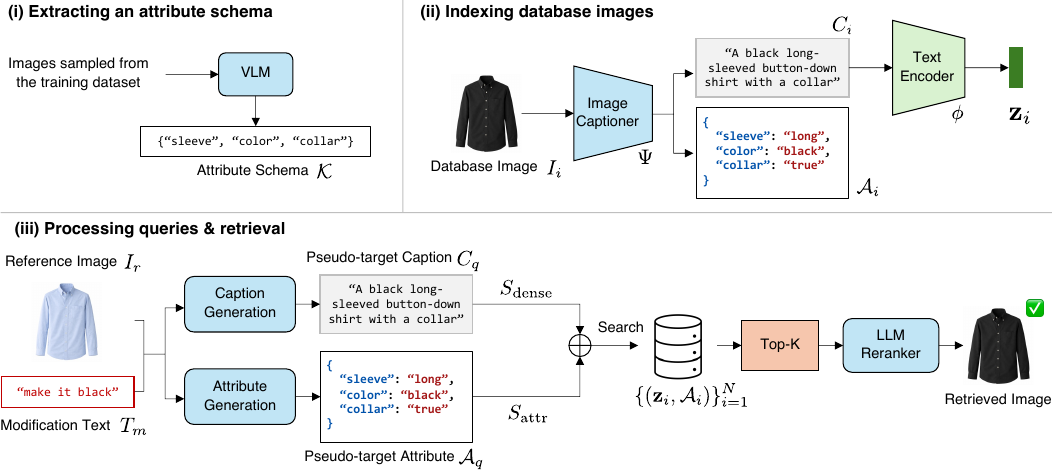}
    \caption{\normalfont Overview of our proposed Vision-Free CIR framework with Attribute-Augmented Hybrid Scoring and LLM-Based Reranking$^1$}
    \label{fig:pipeline}
\end{figure*}

\section{Introduction}
Composed Image Retrieval (CIR) is a task to retrieve a target image using a composite query. This query combines a reference image and a modification text~\cite{vo2019composing}. For instance, an image of a red dress and the text ``change to black'' retrieves an image of a black dress. The expansion of social media and e-commerce drives current interest in CIR.
Conventional methods extract visual features using image encoders to retrieve target.

In contrast, ``Vision-Free'' approaches~\cite{ntinou2025vision} recently emerged and offer a compelling alternative in the broader text-to-image field. These methods represent database images as captions generated by a Vision-Language Model (VLM). The system extracts text embeddings from these captions and retrieves target images based on these embeddings. Thus, this approach does not use image encoders. As highlighted by Ntinou et al.~\cite{ntinou2025vision}, this paradigm offers fundamental advantages. 
First, it eliminates the modality gap inherent in dual-encoder architectures (e.g., CLIP) by processing both modalities with a shared single encoder. 
Second, unlike visual encoders in VLMs that often exhibit ``bag-of-words'' behavior~\cite{clipbagofwords}, this approach leverages LLMs' compositional reasoning capabilities, enabling a  precise understanding of complex attribute binding and logic. 
Finally, it ensures interpretability. Unlike visual embeddings, which act as black boxes, representing images as human-readable text allows for transparent analysis of the rationale behind retrieval results. Inspired by this paradigm shift and the rapid evolution of VLMs, one may consider that this Vision-Free approach is also applicable to a more complex task, CIR.

However, applying this Vision-Free approach to CIR degrades performance. Compared to visual embeddings, text descriptions are inherently lossy and often fail to capture fine-grained visual details such as subtle color distinctions or texture patterns. This limitation is particularly critical in CIR, where modification instructions often specify precise visual attributes (e.g., ``darker blue''). Without explicit mechanisms to preserve such details, Vision-Free CIR significantly underperforms vision-based methods.

To develop the Vision-Free paradigm for CIR, we introduce two practical techniques. First, Attribute-Augmented Hybrid Scoring extracts and matches visual attributes to compensate for missing details. Second, LLM-Based Reranking leverages LLM reasoning to verify semantic consistency.
Our main contributions are summarized as follows:
\begin{itemize}
    \item We propose the first Vision-Free framework for CIR, eliminating the dependency on visual encoders by representing both queries and database images as text.
    \item We introduce two novel techniques to address the information loss in text-only representations: Attribute-Augmented Hybrid Scoring for explicit visual constraint verification, and LLM-Based Reranking for semantic consistency refinement.
    \item Through extensive experiments, we validate that both proposed mechanisms contribute to accuracy gains. Our results demonstrate that Vision-Free CIR outperforms existing approaches on open-domain retrieval (CIRR: 44.04\% R@1).
\end{itemize}

This work aligns with the SynthIR workshop's focus on synthetic content in retrieval ecosystems~\cite{SynthIR2026}, as our framework relies on LLM-generated captions and attributes as retrieval intermediates.

\section{Related Work}
\noindent\textbf{Composed Image Retrieval (CIR).}
CIR has been primarily addressed through supervised learning using triplet datasets~\cite{vo2019composing, val, combiner, aacl}. However, with the advent of powerful VLMs such as CLIP~\cite{clip} and BLIP~\cite{blip}, research focus has shifted toward Zero-shot CIR, which requires no task-specific training data. A representative approach in this domain is Textual Inversion, which maps reference images to pseudo-text tokens to compose them with modification text~\cite{pic2word, searle, fti, contexti2w, gu2024language}. More recently, approaches leveraging the advanced captioning capabilities of VLMs and LLMs have been proposed~\cite{CIReVL, LDRE, GRB}. These methods convert reference images into detailed captions to compose queries entirely within the natural language space. However, while these methods innovate in query composition, they generally rely on image embeddings extracted by visual encoders (e.g., CLIP) to represent and index the database images.

\vspace{0.5em}
\noindent\textbf{Vision-Free Retrieval.}
Recent studies have challenged the necessity of visual encoders in image retrieval systems. Ntinou et al.~\cite{ntinou2025vision} proposed a ``Vision-Free'' architecture that shifts the retrieval paradigm from text-to-image to text-to-text. Instead of indexing visual embeddings, they utilize VLMs to generate structured textual descriptions for database images, enabling retrieval using a single text encoder. This approach has demonstrated state-of-the-art performance on standard retrieval benchmarks. 

Inspired by this success, we extend the Vision-Free paradigm to the more challenging task of CIR, investigating whether textual representations can sufficiently capture the visual nuances required for processing reference images and modification instructions.

\section{Method}
Figure~\ref{fig:pipeline} illustrates our approach for Vision-Free CIR. We first introduce the Vision-Free CIR framework in Sec.~\ref{sec:vision-free-cir-framework}. This framework converts the multimodal task into a text-to-text matching problem. To address the performance degradation caused by this conversion, we propose two practical techniques: 
(1) Attribute-Augmented Hybrid Scoring to explicitly capture visual details (Sec.~\ref{sec:attribute-augmented-hybrid-scoring}) , and 
(2) LLM-Based Reranking to refine the ranking based on semantic consistency between queries and candidates (Sec.~\ref{sec:llm-reranking}). These components are designed as modular extensions. They can be integrated into the Vision-Free framework either individually or in combination, allowing for flexible trade-offs between accuracy and computational cost.

\subsection{Vision-Free CIR Framework}
\label{sec:vision-free-cir-framework}
Let $\mathcal{I}$ be the set of all possible images, and $\mathcal{T}$ be the set of all possible strings.
Consider a database $\mathcal{D} = \{I_i\}_{i=1}^N$ containing $N$ images, where $I_i \in \mathcal{I}$. A CIR instance forms a triplet $(I_r, T_m, I_t)$. Here, $I_r \in \mathcal{I}$ denotes the reference image, and $T_m \in \mathcal{T}$ represents the modification text. The system combines these query inputs to retrieve the ground truth target $I_t \in \mathcal{D}$.
In our Vision-Free framework, we define three key mapping functions. 
The captioner $\Psi: \mathcal{I} \to \mathcal{T}$ converts an image to its descriptive text.
The editor $\Gamma: \mathcal{T} \times \mathcal{T} \to \mathcal{T}$ performs textual modification based on two input strings. The text encoder $\phi: \mathcal{T} \to \mathbb{R}^d$ projects text into a $d$-dimensional embedding space.

First, we represent the entire image database in the text domain. For each image $I_i \in \mathcal{D}$, we generate a caption $C_i = \Psi(I_i)$ and pre-compute its text embedding $\mathbf{z}_i = \phi(C_i)$.
Given a query pair $(I_r, T_m)$, the system first describes the content of $I_r$ as $C_r = \Psi(I_r)$. Next, the editor $\Gamma$ synthesizes a \textit{pseudo-target caption} $C_q = \Gamma(C_r, T_m)$.
Specifically, the editor $\Gamma$ is instructed to rewrite $C_r$ by applying the visual changes specified in the modification text $T_m$, thereby representing the expected visual appearance of the target image.
We compute the cosine similarity between the pseudo-target caption embedding $\phi(C_q)$ and the database embedding $\mathbf{z}_i$. We define this value as the dense score $S_{\mathrm{dense}}(I_i)$:
\begin{equation}
S_{\mathrm{dense}}(I_i) = \cos(\phi(C_q), \mathbf{z}_i)
\label{eq:dense_score}
\end{equation}
The system retrieves the target image $I^*$ by maximizing this score.
\begin{equation}
I^* = \argmax_{I_i \in \mathcal{D}} S_{\mathrm{dense}}(I_i)
\label{eq:dense_retrieval}
\end{equation}

\subsection{Attribute-Augmented Hybrid Scoring}
\label{sec:attribute-augmented-hybrid-scoring}
While text embedding-based retrieval excels at capturing global semantic similarity, it often struggles with compositional reasoning, such as distinguishing ``a white shirt with a green logo'' from ``a green shirt with a white logo''. To mitigate this, we introduce an attribute-augmented scoring mechanism that explicitly aligns visual attributes. Our approach consists of two phases: (1) automated schema discovery, (2) attribute-augmented retrieval.

\vspace{0.5em}
\noindent\textbf{Automated Schema Discovery.}
To ensure generalizability across different datasets without human intervention, we automatically establish an attribute schema $\mathcal{K} \subset \mathcal{T}$. This set $\mathcal{K}$ comprises attribute keys (e.g., $\mathcal{K} = \{\texttt{sleeve}, \texttt{color}, \texttt{collar}\}$ in Figure~\ref{fig:pipeline} (i)). We utilize a training set $\mathcal{D}_{\mathrm{train}}$ separate from the retrieval database $\mathcal{D}$. Specifically, we prompt an off-the-shelf VLM to identify salient categories from a random subset of $\mathcal{D}_{\mathrm{train}}$. This process yields a tailored schema $\mathcal{K}$ derived purely from the data distribution.

Based on $\mathcal{K}$, we use the captioner $\Psi$ to extract an attribute set $\mathcal{A}_i = \{a_i^k\}_{k \in \mathcal{K}}$ for every image $I_i$ in the database, where $a_i^k \in \mathcal{T}$ denotes the value for attribute $k$.
For instance, in Figure~\ref{fig:pipeline} (ii), the extracted values include $a_i^\texttt{sleeve} = \text{"long"}$, $a_i^\texttt{color} = \text{"black"}$, and $a_i^\texttt{collar} = \text{"true"}$.

\vspace{0.5em}
\noindent\textbf{Attribute-Augmented Retrieval.}
Generating \textit{pseudo-target attributes} follows the same two-step logic as generating pseudo-target captions. 
First, we employ the captioner $\Psi$ to extract the reference attribute set based on the schema: $\mathcal{A}_r = \Psi(I_r; \mathcal{K})$. Next, the editor $\Gamma$ predicts the pseudo-target attributes: $\mathcal{A}_q = \Gamma(\mathcal{A}_r, T_m)$. Here, $\Gamma$ is instructed to update only the attribute values modified by $T_m$ while preserving the unrelated ones (e.g., changing \texttt{color} from ``blue'' to ``black'' while keeping \texttt{sleeve}).

We compute the attribute score $S_{\mathrm{attr}}$ by comparing the predicted query attributes $\mathcal{A}_q$ and the database attributes $\mathcal{A}_i$. We employ an exact matching strategy for each attribute. Specifically, we define the similarity function that returns $1.0$ if the predicted value matches the database value, and $0.0$ otherwise. The overall attribute score is calculated as a weighted average:
\begin{equation}
    S_{\mathrm{attr}}(I_i) = \frac{\sum_{k \in \mathcal{K}} w_k \cdot \mathbb{I}(a_q^k = a_i^k)}{\sum_{k \in \mathcal{K}} w_k}
\end{equation}
where $\mathbb{I}(\cdot)$ denotes the indicator function, and $w_k$ represents the weight assigned to attribute $k$. Finally, to balance semantic understanding with strict attribute alignment, we combine the dense retrieval score $S_{\mathrm{dense}}$ (Eq.~\ref{eq:dense_score}) and the attribute score:
\begin{equation}
    S_{\mathrm{final}}(I_i) = \alpha S_{\mathrm{dense}}(I_i) + (1 - \alpha) S_{\mathrm{attr}}(I_i)
    \label{eq:hybrid-score}
\end{equation}
We determine the optimal values for the attribute weights $w_k$ and the balancing parameter $\alpha$ via grid search on a held-out query subset sampled from the training split.

\subsection{LLM-Based Reranking}
\label{sec:llm-reranking}
To accurately capture the semantic relationship between the query and the candidate images, we introduce an LLM-Based Reranking stage. 
This module takes a set of top-$K$ candidates $\mathcal{D}_{\mathrm{top}} \subset \mathcal{D}$ retrieved by the initial scoring stage and reorders them using an LLM.

To perform reranking, we adopt a pointwise scoring approach where the LLM evaluates the combined inputs of the reference image caption, the modification text, and the candidate caption: $(C_r, T_m, C_i)$. 
The model is instructed to output a scalar score in $[0, 10]$ quantifying how well the candidate $C_i$ corresponds to the state of the reference $C_r$ modified by $T_m$.
By directly reasoning over the raw instruction $T_m$, this approach mitigates the information loss inherent in generating an intermediate pseudo-target caption $C_q$.
Finally, the candidates in $\mathcal{D}_{\mathrm{top}}$ are reranked based on these scores to produce the final retrieval result. 

It is worth noting that this explicit LLM-based reranking is a unique advantage of our architecture. Because standard CIR models operate in continuous embedding spaces, they cannot structurally append an off-the-shelf text-only LLM as a reranker without additional cross-modal training. Our Vision-Free pipeline inherently unlocks this capability by homogenizing all inputs into text.

\section{Experiments}

\subsection{Experimental Setup}
\noindent\textbf{Datasets and Metrics.}
We evaluate our method on two standard CIR benchmarks: CIRR~\cite{cirr} and FashionIQ~\cite{fashioniq}.
CIRR is an open-domain dataset containing diverse real-life images, serving as a benchmark for general retrieval capabilities.
In contrast, FashionIQ is a domain-specific dataset specialized in fashion items.
Following prior works, we perform evaluations on the validation split and adopt the Recall@K ($R@K$) metric.

\vspace{0.5em}
\noindent\textbf{Implementation Details.}
We utilize GPT-4.1~\cite{openai2026gpt41} for the captioner ($\Psi$) and the editor ($\Gamma$). For the text encoder ($\phi$), we employ the text-embedding-3-large model~\cite{openai2024embedding}.
For the automated schema discovery, we sampled 100 training images from the training split (per category for FashionIQ, and globally for CIRR). We then prompted the GPT-4.1 to identify the top-4 salient attributes tailored to each domain. For instance, the system selected fine-grained details like \texttt{sleeve\us length} for FashionIQ, while identifying broader concepts like \texttt{number\us of\us subjects} for CIRR.
To determine the hyperparameters (Eq.~\ref{eq:hybrid-score}), we performed a grid search on a held-out subset of 100 queries sampled from the training split.
For LLM-Based Reranking, we utilize GPT-4.1 to serve as the relevance judge.
To balance computational efficiency with retrieval performance, we limit the reranking process to the top-$K=50$ candidates retrieved by the initial scoring. The model is instructed to output a scalar relevance score in the range of $[0, 10]$ for each candidate. 

We compare our method against three zero-shot approaches: SEARLE~\cite{searle}, CIReVL~\cite{CIReVL}, and LinCIR~\cite{gu2024language}. Note that we evaluate these baselines without LLM reranking, as their reliance on visual embeddings makes the direct integration of text-only LLMs structurally infeasible.

\subsection{Results}
\noindent\textbf{Main Results.} Table~\ref{tab:main_results} presents the quantitative comparison with existing zero-shot methods. On the CIRR dataset, our method achieves strong performance, recording 44.04\% in R@1. This score surpasses the existing competitive method, LinCIR (ViT-G backbone), by +8.79\%.
Our approach leverages the reasoning power of LLMs to operate entirely within the text domain. This suggests that textual representations outperform visual embeddings in open-domain retrieval requiring logical consistency.

On the FashionIQ dataset, our method outperforms baselines utilizing ViT-L backbones in terms of R@10 driven by LLM-Based Reranking. However, we observe a distinct performance gap compared to SOTA methods, particularly in R@50. This gap stems from the fundamental difficulty of representing domain-specific visual details without image encoders. However, the relatively high R@10 score highlights the efficacy of our LLM-Based Reranking. 

\begin{figure}[t]
    \centering
    \includegraphics[width=\linewidth]{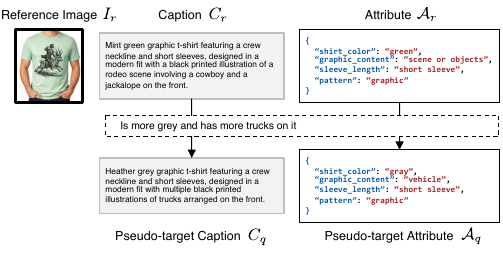}
    \caption{\normalfont Example of caption and attribute generation$^1$.}
    \label{fig:modification_example}
\end{figure}

Figure~\ref{fig:modification_example} illustrates the generation of reference and pseudo-target metadata. We observe that the reference caption and attributes accurately describe the original image. Furthermore, the pseudo-target representations correctly incorporate the modification text, as seen in the \texttt{shirt\us color} attribute shifting from ``green'' to ``gray''.

\begin{table}[t]
\centering
\caption{\normalfont Zero-shot CIR results on CIRR and FashionIQ datasets. For FashionIQ, we report the average scores across all categories.}
\label{tab:main_results}
\small
\setlength{\tabcolsep}{3pt}

\begin{tabular}{@{}l | c | ccc | cc@{}}
\toprule
\multirow{2}{*}{\textbf{Method}} & \multirow{2}{*}{\textbf{Backbone}} & \multicolumn{3}{c|}{\textbf{CIRR}} & \multicolumn{2}{c}{\textbf{FashionIQ (Avg)}} \\
 & & R@1 & R@5 & R@10 & R@10 & R@50 \\
\midrule
SEARLE~\cite{searle} & \multirow{3}{*}{ViT-L} & 24.24 & 52.48 & 66.29 & 25.56 & 46.23 \\
CIReVL~\cite{CIReVL} & & 24.55 & 52.31 & 64.92 & 28.29 & 49.35 \\
LinCIR~\cite{gu2024language} & & 25.04 & 53.25 & 66.68 & 26.28 & 46.49 \\
\midrule
SEARLE & \multirow{3}{*}{ViT-G} & 34.80 & 64.07 & 75.11 & 34.81 & 55.71 \\
CIReVL & & 34.65 & 64.29 & 75.06 & 28.55 & 48.57 \\
LinCIR & & 35.25 & 64.72 & 76.05 & \textbf{45.11} & \textbf{65.69} \\
\midrule

\textbf{Ours} & Text-Emb-3 & \textbf{44.04} & \textbf{70.53} & \textbf{77.66} & 34.59 & 44.93 \\
\bottomrule
\end{tabular}
\end{table}

\begin{table}[t]
\centering
\caption{\normalfont Ablation study of the proposed components.}
\label{tab:ablation}
\small
\setlength{\tabcolsep}{2pt}

\begin{tabular}{@{}l ccc cc@{}}
\toprule
\multirow{2}{*}{\textbf{Configuration}} & \multicolumn{3}{c}{\textbf{CIRR}} & \multicolumn{2}{c}{\textbf{FashionIQ (Avg)}} \\
\cmidrule(lr){2-4} \cmidrule(lr){5-6}
 & R@1 & R@5 & R@10 & R@10 & R@50 \\
\midrule
\textbf{1. Baseline}                     & 12.20 & 48.18 & 61.53 & 23.86 & 40.40 \\
\textbf{2. 1 + Attribute Scoring}        & 14.83 & 50.86 & 64.07 & 27.04 & 44.93 \\
\textbf{3. 1 + LLM Reranking}            & 43.56 & 70.12 & 77.58 & 32.51 & 40.33 \\
\textbf{4. Full (2 + LLM Reranking)}     & 44.04 & 70.53 & 77.66 & 34.59 & 44.93 \\
\bottomrule
\end{tabular}
\end{table}

\vspace{0.5em}
\noindent\textbf{Impact of Attribute-Augmented Hybrid Scoring.}
\begin{figure}[t]
    \centering
    \includegraphics[width=\linewidth]{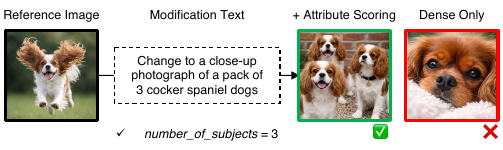}
    \caption{\normalfont Qualitative analysis for attribute-augmentation$^1$.}
    \label{fig:attribute_success_example}
\end{figure}
\footnotetext[1]{Due to licensing restrictions, we show generated images that are visually similar to those in the original datasets.}
Attribute scoring (Table~\ref{tab:ablation}, Row 2) consistently improves the baseline. Crucially, it raises the theoretical ceiling for the subsequent LLM reranking. Since reranking only evaluates the top-50 candidates, the initial R@50 strictly bounds final performance. By boosting FashionIQ's R@50 from 40.40\% to 44.93\%, attribute scoring acts as an indispensable coarse filter, keeping correct targets in the candidate pool.

Qualitatively, the benefit of Attribute-Augmented Hybrid Scoring is evident in queries requiring strict constraints. Figure~\ref{fig:attribute_success_example} shows the example query from the CIRR dataset. The baseline dense-only model captured the species and camera angle but ignored the numerical constraint, retrieving a single dog. In contrast, our model predicted the \texttt{number\us of\us subjects} as ``3''. This explicit match boosted the rank of the correct target image, leading to a successful retrieval.

\vspace{0.5em}
\noindent\textbf{Impact of LLM-Based Reranking.}
\begin{figure}[t]
    \centering
    \includegraphics[width=\linewidth]{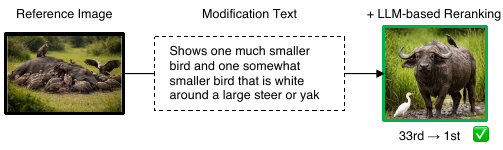}
    \caption{\normalfont Qualitative analysis for LLM-Based Reranking$^1$.}
    \label{fig:reranking_success_example}
\end{figure}
The application of LLM-Based Reranking (Table~\ref{tab:ablation}, Row 3) following Attribute-Augmented Hybrid Scoring significantly improves performance across both datasets. In FashionIQ, R@10 improves from 27.04\% to 34.59\%, while in CIRR, R@1 increases from 14.83\% to 44.04\%. These results indicate that the reranker effectively identifies correct targets initially ranked lower (within the top-50) and promotes them to the top positions.

Figure~\ref{fig:reranking_success_example} shows a qualitative example. In this case, the modification text involves complex constraints regarding multiple objects. The LLM correctly interprets these intricate relationships, successfully boosting the correct target from rank 33 to 1.

\section{Conclusion}
In this paper, we proposed a Vision-Free CIR framework that eliminates visual encoders, enhancing accuracy via Attribute-Augmented Hybrid Scoring and LLM-Based Reranking. Our method outperformed existing zero-shot approaches on the open-domain CIRR dataset. Conversely, results on FashionIQ highlight the challenge of capturing fine-grained visual details without image embeddings.

\bibliography{references}

\end{document}